\documentclass{article}
\usepackage{spconf,amsmath,graphicx}
\usepackage{bbm}

\usepackage{times}
\usepackage{color}
\usepackage{amsmath}
\usepackage{graphicx,psfrag,epsf}
\usepackage{enumerate}
\usepackage{url}
\usepackage{amssymb,amsfonts,mathtools}
\usepackage{bm}
\usepackage{color, colortbl}
\usepackage{algorithm}
\usepackage{subcaption}
\usepackage{mathtools}

\usepackage{balance,cite}
\usepackage{multirow}
\usepackage{bbm}

\DeclareMathOperator*{\argmin}{arg\,min}

\DeclarePairedDelimiterX{\norm}[1]{\lVert}{\rVert}{#1}
\DeclarePairedDelimiterX{\bnorm}[1]{\biggl\lVert}{\biggr\rVert}{#1}
\DeclarePairedDelimiterX{\abs}[1]{\lvert}{\rvert}{#1}

\def\R{\mathbb{R}}
\def\sign{\textrm{sign}}
\def\E{\mathbb{E}}
\def\P{\mathbb{P}}
\def\de{\overset{\Delta}{=}} 
\def\F{\mathcal{F}}
\def\X{\mathcal{X}}
\def\Y{\mathcal{Y}}

\newcounter{custom}

\title{Federated Learning Challenges and Opportunities: An Outlook}
\name{
Jie Ding, Eric Tramel, Anit Kumar Sahu, Shuang Wu, Salman Avestimehr, Tao Zhang
}
\address{Alexa AI, Amazon}

\begin{document}
%
\maketitle
\begin{abstract}

Federated learning (FL) has been developed as a promising framework to leverage the resources of edge devices, enhance customers' privacy, comply with regulations, and reduce development costs. Although many methods and applications have been developed for FL, several critical challenges for practical FL systems remain unaddressed. This paper provides an outlook on FL development as part of the ICASSP 2022 special session entitled ``Frontiers of Federated Learning: Applications, Challenges, and Opportunities.'' The outlook is categorized into five emerging directions of FL, namely algorithm foundation, personalization, hardware and security constraints, lifelong learning, and nonstandard data. Our unique perspectives are backed by practical observations from large-scale federated systems for edge devices.

\end{abstract}
%

\vspace{-0.1in}
\section{Introduction} \label{sec_intro}
\vspace{-0.1in}

\begin{figure}[b]
  \centering
  \vspace{-0.1in}
  \includegraphics[width=.9\linewidth]{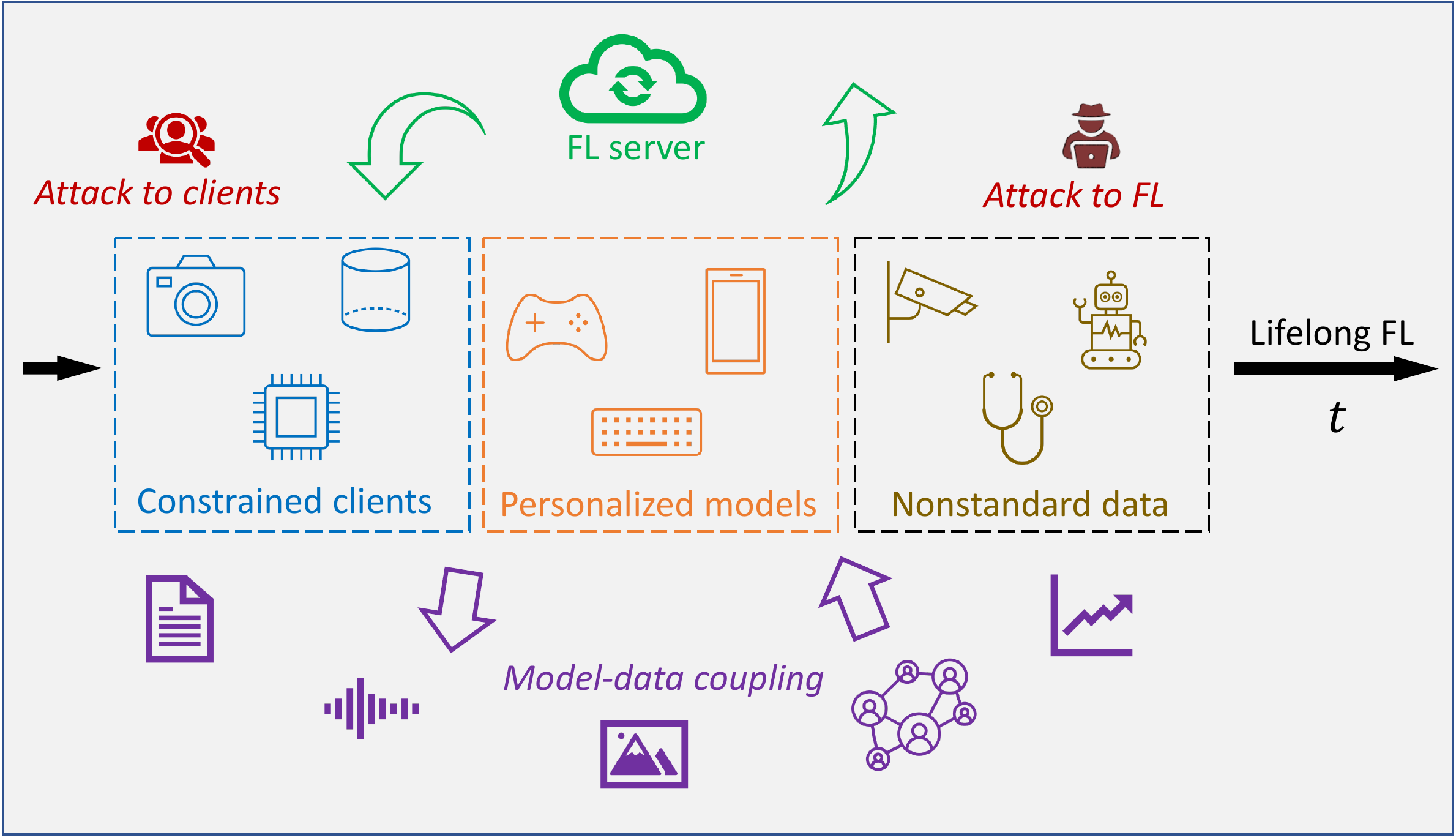}  
  \vspace{-0.1in}
  \caption{Illustration of an FL system in our outlook.}
  \label{fig_overview}
\vspace{-0.1in}
\end{figure}

Federated learning~\cite{shokri2015privacy,konevcny2016federated} is a popular distributed learning framework developed for edge devices. It allows the private data to stay locally while leveraging large-scale computation from edge devices. Its main idea is to learn a joint model by alternating the following in each so-called federated, or communication, round: 1) a server pushes a model to clients, who will then perform multiple local updates, and 2) the server aggregates models from a subset of clients. 
The design of practical FL systems is highly nontrivial since FL often involves millions of devices, unknown heterogeneity from various cohorts, limited on-device capacity, evolving data distributions, and partially labeled data. Inspired by our practical observations, we will list some critical challenges in the following five sections (demonstrated in Figure~\ref{fig_overview}). 

\vspace{-0.1in}
\section{Algorithmic fundamentals of FL}
\label{sec_fund}
\vspace{-0.1in}



\textbf{FL goals}. 
There are two general goals of supervised learning, namely prediction, which aims to learn a model with a small out-sample prediction risk, and detection, which will maximize the statistical power under a fixed false-detection rate. A principled FL design needs to consider unique perspectives that depend on the particular goal. We take the binary classification task as an illustrating example. Similar observations apply to more complicated tasks. Let the training data $(Y_i,X_i)$, $i=1,\ldots,n$, and test data $(Y,X)$ be IID random variables with values in $\R^d \times \{1,-1\}$.

1) \textit{Prediction}. For the prediction purpose, one often looks for a classifier $C$ 
such that the risk $\P(C(X) \neq Y)$ is small. It is known that the optimal classifier is $C_*: x \mapsto \sign(f_*(x)) \in \{1,-1\}$ 
and $f_*(x) = \E(Y=1 \mid x) - 1/2$. 
To train a classifier $\hat{C}_n$, a general approach is to learn $\hat{f}_n: \R^d \rightarrow \R$ and then let $\hat{C}_n(x) \de \sign(\hat{f}_n(x))$.
Specifically, we operate the estimation from an (often) infinite-dimensional function space with only finite sample through sieve estimators, to trade-off between approximation errors and estimation variance, e.g., the empirical risk minimization
\begin{align}
    \hat{f}_n = \argmin_{f \in \F} \frac{1}{n} \sum_{i=1}^n L(Y_i, f(X_i)) + \lambda_n \Omega(f), \label{eq1}
\end{align}
where $\F$ is a function class, 
$\Omega$ is a regularization functional, and $L$ is a loss function.
The fundamental problem is to choose a proper modeling vehicle to attain a rate-optimal $\hat{f}_n$ regarding the prediction risk. 
In the centralized setting, asymptotically optimal rules for rather general models have been studied 
(see~\cite{DingOverview,DingLOL} and the references therein).
In FL settings, however, there may not exist a way to equivalently operate (\ref{eq1}), not to mention optimality, for general $\F$, $L$, $\Omega$, and $\lambda_n$ (except for finite-dimensional $\F$ and fixed-shape parameters). 
Thus, it is important to develop a deeper understanding of whether (\ref{eq1}) admits an FL-based solution.

2) \textit{Detection}.
The other goal is to develop a powerful test of $H_0: y=-1$ against $H_1: y=1$ given $x$. This detection problem is critical for many applications, e.g., keyword detection, anomaly detection, and drug discovery. Given a fixed false-detection rate $\P(f(x)>0 \mid y=-1)$, we want to minimize the false-rejection rate $\P(f(x)<0 \mid y=1)$. The optimal decision rule is then given by the Neyman-Pearson Lemma and represented as a ROC curve. 
Though both goals require estimating a function $f$, they are fundamentally different. An optimal decision rule for the prediction goal is often not suitable for detection. An example concerns a highly imbalanced dataset, say $\P(Y=1) \gg \P(Y=-1)$. 
In that case, the detection rule only depends on $X \mid Y=-1$ and $X \mid Y=1$, which is not affected by $\P(Y=1)$, while the prediction rule uses $Y \mid X$, which is sensitive to the observed frequency of $Y=1$. Consequently, a larger imbalance tends to produce a model that favors the majority class and leads to varying detection thresholds (if using the probability of $Y \mid X$) under a fixed false-detection rate.
An example is the detection of occurrences of a keyword in a long data stream.
Although the conflict and reconciliation of prediction- and detection-oriented modeling have been studied for centralized settings~\cite{DingKinetic,DingOverview}, little is known in FL settings. A particular question is whether an FL model of good prediction performance, trained from (\ref{eq1}), will inevitably sacrifice detection power.


\textbf{Limits of communication efficiency (in rounds)}.
Empirical evidence shows that the standard FL that averages over suitably many local updates for communication efficiency can still converge to a desirable model. Meanwhile, overly aggressive local updates are likely to harm the performance due to the heterogeneity of finite data across clients.  
The fundamental problem is understanding lower limits of the number of communication rounds needed to converge to a model that performs similarly to centralized training. 
Consider two extremes of FL training. One is to let each client perform one step of local update, which corresponds to a centralized SGD training~\cite{mcmahan2017communication}. Under reasonable conditions, 
the training directly corresponds to the optimization of (\ref{eq1}) but is not communication efficient.
The other extreme is to let each client perform sufficiently many local steps until its convergence, and then the server aggregates only once. For example, suppose that two clients hold $n_1$ and $n_2$ different data, and they individually solve (\ref{eq1}) to obtain $\hat{f}_{n_1}$ and $\hat{f}_{n_2}$, respectively. We say the problem (\ref{eq1}) meets \textit{$\alpha$-divisibility} if the risk (after subtracting the Bayes optimal risk), denoted by $R$, satisfies $R(w_1 \hat{f}_{n_1} + (1-w_1) \hat{f}_{n_2}) \leq \alpha R(\hat{f}_{n})$ for all $w_1 = n_1/n$. Clearly, a smaller divisibility $\alpha$ implies less risk degradation in FL. For a constant $\alpha$, the order of the risk rate is maintained with only one round of communication, representing the most communication efficiency.  It is easy to show that $\alpha$ is asymptotically fixed for parametric $\F$ and some standard regularity conditions. 
In general, however, $\alpha$ may not be small, depending on many factors such as the function class, regularization, and loss types. The limits of communication rounds needed for rate-optimality in FL (if it exists) have yet to be studied.

\vspace{-0.1in}
\section{Personalization of FL}
\label{sec_person}
\vspace{-0.1in}

Many FL applications aim to improve the clients of heterogeneous data. Examples are personalized voice assistants, precision medicine, cohort-specific autonomous driving. Unlike the conventional FL that aims to train one model, 
the goal of personalized FL is to learn a collection of client-specific models that reduce test errors beyond what is possible with a single global model. 

\textbf{Nature of personalization}. 
Existing work on personalized FL often derives algorithms based on a heuristic optimization formulation, which aims to regularize the discrepancy between local parameters and global parameters. We consider the following generic formulation to operate the personalized FL from the server's perspective.
\begin{align}
	\min_{f_1,\ldots,f_M \in \F} \sum_{m=1}^M G_{m,n_m}(f_m) + \lambda_n \Omega_n(f_1,\ldots,f_M), \label{eq_stat} 
\end{align}
where $G_{m,n_m}$ denotes client $m$'s local loss, $\F$ is the common function class, and $\Omega_n$ is a regularization term.
Without $\Omega_n$, it is equivalent to optimizing $M$ personal objectives separately.  
The above formulation raises two concerns. First, it is unclear how to choose $\Omega$. Even if one heuristically specifies a regularization, \eqref{eq_stat} does not tell what an upper-bound baseline is for a client's local performance by participating in FL. 
There are two direct lower-bound baselines of personalized FL, namely, each client performs local training without FL, and all clients participate in conventional FL training. Understanding the gap between lower and upper limits is practically essential and theoretically intriguing. The second concern is whether the current heuristic approaches can indeed leverage the server-side global model as an intermediary to exchange helpful information across clients. Designing a personalized FL requires calibrating the tradeoff between a client's local training and inheritance from the aggregated model. Sieve estimators beyond the empirical risk minimization (\ref{eq_stat}) have yet to be developed. 
It is also interesting to establish connections between personalized FL with other frameworks such as assisted learning~\cite{DingAssist}, meta learning~\cite{finn2017model}, multi-task learning~\cite{evgeniou2004regularized}, and knowledge distillation~\cite{hinton2015distilling}.


\textbf{Connections of personalization, fairness, and robustness}. 
Suppose that there are different cohorts of users to which personalized FL is applied. It is foreseeable that the model trained for one cohort may not be systematically biased due to the influence of other cohorts. Thus, better personalization can be associated with improved fairness (in a proper sense). Meanwhile, the global model trained from personalized FL is expected to be more robust to the heterogeneity of particular clients, which leads to enhanced robustness. Thus, fundamental connections between personalization, fairness, and robustness in the FL setting deserve further research.

\textbf{Is a personalized model sufficiently good?}
There are two kinds of quality assessment of a model in general. One is based on a utility (say accuracy), and the other is based on the systematic discrepancy with the underlying data generating process. For example, if $Y$ is almost independent of $X$, the optimal predictive model has accuracy near one-half. Thus, a model with unsatisfactory accuracy does not necessarily mean we could further improve it. Instead, it depends on the nature of the task. The assessment of systematic defects of a model is often studied in the context of goodness-of-fit tests~\cite{DingBAGofT} and is an underexplored problem in the personalized FL setting. With more studies along this direction, we can understand whether an undesirable prediction is due to the incompleteness of FL designs or fundamental limitations of the data and task.

\vspace{-0.15in}
\section{Constrained FL: Memory, Computation, Security, and privacy}
\vspace{-0.15in}

The predictive performance of a general FL system is determined by the approximation error (bias), estimation error (variance), and optimization error (gap with the intended minimum), which are further subject to the specified model, the nature of $Y \mid X$, the optimization solver, and the way data are distributed. While a deeper understanding of their tradeoffs in various settings is yet to be studied, there are additional practical constraints that add to the complication of FL. 

\textbf{Memory and computation constraints}.
A practical FL system for edge devices has to accommodate on-device hardware capacity. We discuss two tradeoff factors. First, under memory constraints, each on-device model needs to be small in size. This poses a methodological challenge for the server to leverage a large function class for large data (in hindsight). Promising directions include the use of knowledge transfer~\cite{FedGKT} and knowledge distillation~\cite{hinton2015distilling}, heterogeneous on-device models with neural parameter sharing~\cite{DingHeteroFL}, or neural architecture search~\cite{he2020fednas}. 
Second, under computation constraints, devices may perform only a limited number of gradient updates~\cite{FLFieldGuide}. This leads to the challenge of prolonged FL training time, which requires developing an efficient optimization solver on the client-side and an effective aggregation rule on the server-side. Also, lower-energy devices are infrequently activated, leading to a low client-sampling rate and potentially significant fluctuations in the aggregated model.

\textbf{Security and privacy constraints}.
Since FL does not require raw data transmission, such as customers' images and audios, it is naturally suitable for improving data privacy. However, some research has shown that clients' identifiable information can be extracted from gradients~\cite{orekondy2018gradient}. It has motivated privacy-preserving techniques such as differentially private SGD. 
To mitigate the risk of on-device data being compromised, one may only keep small data on a device in a period, causing additional challenges in FL training (in Section~\ref{sec_life}).
A related but distinct privacy concern is to deploy an already-developed but proprietary model API to the open world. In that case, we need a deeper understanding of tradeoffs between utility and model privacy~\cite{DingIL}. 
Also, during the FL training, there has been a surge of recent algorithms to provide secure aggregation~\cite{so2021turbo,multiRoundSecAgg,beguier2020safer}, where the general idea is to only learn aggregated parameters over cohorts of devices instead of single devices for the server. 
From the adversarial learning perspective, since an FL system is built on numerous participants, it can be vulnerable in situations outside the intended design, e.g., when some participants maneuver the training rule.
It is critical for the FL designer to understand tradeoffs between prediction and resilience against various attacks and quantify an FL system's vulnerability in practice.

\vspace{-0.1in}
\section{Lifelong FL} \label{sec_life}
\vspace{-0.1in}

Conventional FL systems are designed to learn only a pre-specified task using data collected from a fixed channel. Thus, their generalization capability is quite limited for most real-world learning situations where the underlying data and tasks vary over time. For example, the on-device data distribution evolves due to user behavioral changes, or the label sets are updated due to new product releases. A general goal of lifelong FL is to develop systems that can {learn continuously during execution}, {improve performance stably over time}, and {adapt existing models to dynamic environment} without forgetting previous learning. We highlight two critical challenges. 

\textbf{Online updates with single-pass data}. 
During the continuous execution of an FL system, a client often receives new data online instead of holding a static local dataset. Correspondingly, the local training in each round involves both existing and newly observed data, causing extra complications to the convergence of an FL model. Moreover, due to memory or security constraints, a client may only store a small window of online data, so the local training cannot replay historical data. In other words, the training has to be made from a single pass of data, bringing a major challenge for reliable online incremental updates of an FL model.

\textbf{Coupling of model training and data generation}. 
In many applications, new local data are collected through a client-side model (e.g., according to the soft-max values), which tend to be selectively biased towards the most recent inference model. Take the keyword detection problem as an example. Suppose that at time $t$, the FL-updated on-device model is $X \mapsto f_t(X)$. It will then be used to identify new data (denoted by $D_{t+1}(f_t)$) to feed into the FL training at time $t+1$. If another FL system were used, say $f_t$ were replace with $f^{'}_t$, the newly fed data would become different, namely $D_{t+1}(f^{'}_t)$.
A generic formulation is
{\small
\begin{align*}
    \textrm{$D_{t}$ and momentum}  & \xRightarrow[\text{Update}]{\text{FL}} f_t, \,
    \textrm{Inference of $f_t$} & \xRightarrow[\text{Collect}]{\text{Data}} D_{t+1},
\end{align*}
}
where $Y$ may denote a ground-truth label or a machine-generated label (Section~\ref{sec_nonstandard}).
Consequently, the training data cannot be treated as IID, and the test data based on predictive validation~\cite{DingOverview} tends to favor the most recently trained model.
These cause practical hazards to evaluate (and thus improve) FL performance, at least in the following two aspects.

1) \textit{Data distribution shift}. The data fed into the FL system at different rounds are Markovian, so any performance evaluation of FL based on historical data may not be reliable. 

2) \textit{Model procedure selection for large-scale dynamical systems}. Model procedure selection in the broad sense refers to the selection of learning models, tuning parameters, and evaluation tools as a part of a modeling procedure~\cite{DingOverview}. It is the key to lifelong FL, as we need to develop a sequence of models over time instead of a single model. Due to {model-data coupling}, standard predictive validation or A/B testing may not apply. We need new methods for efficient deployment and evaluation of multiple FL procedures simultaneously. 
 
\textbf{Rethinking of Batch Normalization (BN) layers to mitigate catastrophic forgetting}.
Catastrophic forgetting means that adaptions to new data or tasks tend to degrade the test performance on the original data or task domain.
This is particularly a concern for lifelong FL because the clients often have heterogeneous data, and their model aggregation can easily introduce large noises and lose useful information.
A potential mitigation of the curse of catastrophic forgetting is to rethink the role of BN layers~\cite{ioffe2015batch} in FL to overcoming data heterogeneity across clients.
Recall that during the training phase, each BN layer standardizes a data batch $x_b$ by
$
    \tilde{x}_b = (\sigma^2+\epsilon)^{-1/2} (x_b-\mu) \cdot \gamma+\beta, 
$ 
where $\gamma, \beta$ are part of the model parameters and $\epsilon$ is a small constant. Here, $\mu, \sigma^2$ are running statistics being tracked with momentum during the training, as a default BN option in popular frameworks that is often taken for granted. 
However, it is unclear how to specify $\mu, \sigma^2$ during training and prediction in the FL setting. It was recently shown that an adaptation of BN named Static Batch Normalization (sBN)~\cite{DingHeteroFL} can significantly accelerate the convergence and improve the performance of FedAvg~\cite{mcmahan2017communication} compared with BN and other forms of normalization. 
Similar observations were also made in~\cite{andreux2020siloed}.
In FL training with sBN, the affine parameters $\gamma$ and $\beta$ are aggregated as usual, but $\mu$ and $\sigma^2$ are only calculated from local data batch, namely the above $x_b$. After training, the server will query local clients and calculate the global statistics of $\mu$ and $\sigma^2$ for prediction.
An intuition is that re-calibrating the BN statistics (as in sBN) can help data of heterogeneous nature leverage an already-trained feature extractor consisting of BN layers.

\vspace{-0.1in}
\section{Data incompleteness, polarity, and complex dependency} \label{sec_nonstandard}
\vspace{-0.1in}

Conventional FL for supervised learning often assumes that labeled data in the standard form of $(X,Y) \in \X \times \Y$ are readily available to train. Here, $\X$ and $\Y$ denote the data space and label space (e.g., $\R^d$ and $\{1,-1\}$ for the example in Section~\ref{sec_fund}), respectively. However, in many real-world applications, we often see data in nonstandard forms due to resource constraints, collection restrictions, or domain-specific formats. 
We summarize three critical but underexplored challenges regarding partially labeled $X$, partially observed $\Y$, and irregular $X$, respectively. 

\textbf{Data incompleteness}. 
In many FL applications, clients cannot access all the ground-truth labels.
For instance, a medical center wishes to use FL to significantly improve the diagnosis quality without transmitting sensitive data from clinics distributed in rural areas. In contrast, clinics do not have advanced medical resources to label all the data. 
In general, suppose that client $m$ holds $n_m^{\textrm{la}}$ labeled data $(X_m^{\textrm{la}}, Y_m^{\textrm{la}})$ and $n_m^{\textrm{un}}$ unlabeled data $X_m^{\textrm{un}}$, $m=1,\ldots,M$. Admittedly, one could use the labeled data only for FL training. However, when $n_m^{\textrm{un}}$ is larger in orders of magnitudes than $n_m^{\textrm{la}}$,  using the unlabeled data wisely may significantly boost the FL performance. This is a barely studied but worthwhile direction of FL. In the centralized setting, several recent works on semi-supervised learning have already shown the great potential of leveraging unlabeled data. 
Among the partially-labeled FL settings, a practical scenario of particular interest concerns the ``fully-unlabeled clients,'' where $n_m^{\textrm{la}}=0$ for all $m$, but the server may have some labeled data.
Back to the example, a medical center could have a few labeled data due to its rich resources, while distributed clinics have only unlabeled data.
For this challenge, recent work on semi-supervised FL~\cite{DingSemiFL} has shown promising baseline results that with properly designed consistency regularization techniques, one could achieve performance close to the centralized and fully-labeled training, but with unlabeled clients. Its extensions to address the challenges in Sections~\ref{sec_person}-\ref{sec_life} are worth investigating. Also, since the existing work mostly focused on image data, there is much to be explored in audio and text domains. 


\textbf{Data polarity}. We say a set of data $(X_i,Y_i)$, $i=1,\ldots,n$, exhibits \textit{polarity} if $Y_i \in \Y^{\textrm{obs}}$ for some $\Y^{\textrm{obs}}\not\subseteq \Y$ and all $i$. For example, a machine based on natural language only stores data that contains a detected keyword for subsequent recognition, so only positive data are collected but negative ones are dropped. In such cases, the collected $(X,Y)$ does not represent the whole data distribution, potentially causing severe biases in prediction. This challenge is further complicated by issues mentioned in Section~\ref{sec_life}, which limits the use of historical data (even if they are balanced).

\textbf{Complex data dependency}. 
While it is convenient to study IID $X \in \R^d$, in many FL applications, data are collected from time series with inevitable dependency. For example, suppose that $\{Z_t\}_{t=1,2,\ldots}$ is a data sequence, and client $m$ observes $\{Z_{m,t}\}_{t=1,2,\ldots}$, where $Z_t \in \R^k$ and $Z_{m,t}$ is a sub-vector of $Z_{t}$.  For the task of one-step ahead prediction of $Z_t$, client $m$'s labeled data are $(X_{m,t}, Y_{m,t})$, where $X_{m,t} = [Z_{m,t-1},\ldots,Z_{m,t-\ell}]$ and $Y_{m,t} = X_{m,t}$ for a properly chosen lag order $\ell$~\cite{DingSlant}. Though one could still apply (\ref{eq1}), it may not be justified since $X_{m,t}$, $t=1,2,\ldots$ are non-IID or even non-stationary, and the label $Y_{m,t}$ is only a sub-vector of the intended $Y_{t}$. This is further complicated if one considers a spatio-temporal structure where $Z_t$ represent a time-varying graph, e.g.,  transportation data, weather data, and internet traffic data observed from different locations.


\vspace{-0.1in}
\section{Concluding Remarks} \label{sec_intro}
\vspace{-0.1in}

Significant progress has been made in federated learning over the past couple of years. However, as we discussed throughout the paper, many challenging and interesting research problems are still left to be studied. In particular, from algorithmic aspects, the issues of communication rounds, personalization, lack of labels, robustness, and continuation in federated learning are not explored much. Furthermore, from system design perspectives, the challenges of resource constraints at the edge nodes, security, and privacy are widely open areas.
In parallel to algorithmic and system-design challenges, FL research is also advancing rapidly in application domains beyond computer vision~\cite{konevcny2016federated} and natural language processing~\cite{FedNLP}. Examples are graph-structured data~\cite{fedGraphNN} and time-series data that arise in drug discovery, social network, recommendation system, and advertisement domains. 

\clearpage
\balance
\bibliographystyle{IEEEbib}
\bibliography{privacy,J}

\end{document}